# Synplex: A synthetic simulator of highly multiplexed histological images.

Daniel Jiménez-Sánchez[1], Mikel Ariz[1] & Carlos Ortiz-de-Solórzano[1]

[1]*Solid Tumors and Biomarkers Program, IDISNA, and Ciberonc, Center for Applied Medical Research, University of Navarra, 31008, Pamplona, Spain*

**Abstract.** Multiplex tissue immunostaining is a technology of growing relevance as it can capture *in situ* the complex interactions existing between the elements of the tumor microenvironment. The existence and availability of large, annotated image datasets is key for the objective development and benchmarking of bioimage analysis algorithms. Manual annotation of multiplex images, is however, laborious, often impracticable. In this paper, we present **Synplex,** a simulation system able to generate multiplex immunostained *in situ* tissue images based on user-defined parameters. This includes the specification of structural attributes, such as the number of cell phenotypes, the number and level of expression of cellular markers, or the cell morphology. **Synplex** consists of three sequential modules, each being responsible for a separate task: modeling of cellular neighborhoods, modeling of cell phenotypes, and synthesis of realistic cell/tissue textures. **Synplex** flexibility and accuracy are demonstrated qualitatively and quantitatively by generating synthetic tissues that simulate disease paradigms found in the real scenarios. **Synplex** is publicly available for scientific purposes, and we believe it will become a valuable tool for the training and/or validation of multiplex image analysis algorithms.

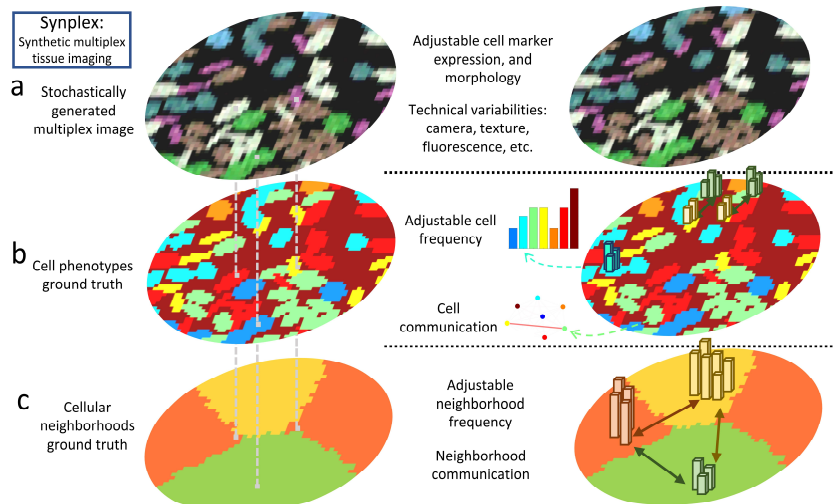

**Figure 1**: Conceptual framework of **Synplex**.

# 1 Introduction

## 1.1 The tumor microenvironment (TME)

Cancer is a tissue-level disease. It can be seen as a *malignant* organ with a complex microenvironment (TME) that includes cancer, transformed cells, immune and stromal cells, blood vessels, and extracellular matrix components [1][2][3]. The organization of these components within the tissue reflects the tumor history, and it is growingly believed to correlated with its response to therapy [4][5]. Therefore, determining which TME elements and relationships are involved in tumor progression and patient prognosis is of the highest clinical utility [6].

Multiplexed *in situ* imaging (MI) technologies allow the simultaneous staining of a high number of proteins expressed in the cells and, thus characterizing multiple cell phenotypes while preserving their spatial location in the tissue [7][8]. In MI, the phenotype of an individual cell can be defined from the abundance, morphology, and turnover of its -labelled- intracellular organelles and cytoskeletal structures [9][10]. Moreover, each identified cell can then be placed within the context of a cell population or tissue.

Understanding the biological events that happen in malignant processes such as tumor formation or progression, is crucial to find new biomarkers and develop new therapies. Multiplex imaging technologies have the potential to help in this task by answering questions such as: which cell phenotypes compose a specific tumor neighborhood (e.g. the peritumoral region)? Do some cell phenotypes interact with each other in a special way? Is this interaction related to the patient's clinical subtype? Is the relative expression of a marker or combination of markers significantly correlated with the patient survival?

## 1.2 Computational tools to quantify the tumor microenvironment

Automated image analysis algorithms can be designed to process vast amounts of multiplex image data while learning complex spatial patterns both at cell and tissue levels, and associate those patterns with the patients' diagnosis or prognosis [11][12]. These methods can be supervised or unsupervised. Supervised methods are used to study the role that a specific cell phenotype or simple interaction between a few cell phenotypes has in the disease. These methods can for instance quantify cells with specific levels of marker expression [13], or measure the colocalization of certain biomarkers in a specific anatomical region [14]. To this end, researches create manual annotations of relevant TME elements, that are fed to Machine Learning (ML) frameworks that automatically learn the underlying tissue pattern [15][16][17]. Unsupervised methods are better suited when the number of markers is high, thus making manual annotations unfeasible. In this case, tissues are analyzed in *discovery* mode, looking for unknown cell phenotypes or interactions between phenotypes that are relevant for the biological process under study [18][19]. To this end, cell segmentation tools extract cell features, e.g., the intensity of the biomarkers, morphological features such as cell size and shape, along with some information of the cell's environment such as the number of neighboring cells and their phenotypes [20]. Then, a tissue-level topological network of interactions can be built and a graph-based approach used to cluster groups of cells into distinct neighborhoods [21]. Those neighborhoods are composed of cell phenotypes, which in certain situations, can be repelled or attracted to each other. Once patients/images are characterized from the information of cell-phenotypes and neighborhood interactions, they could be linked to the task at hand, may it be the classification of tumor types, or the prognosis/diagnosis of patients.

### 1.3  Simulation of the tumor microenvironment.

Regardless of their supervised or unsupervised nature, computational methods must be validated before being used in either research of clinical environments. To this end, platforms that allow an objective, systematic performance evaluation of new computational methods are of great value for the scientific community. These platforms rely on annotated datasets. However, due to the novelty of the technology and the complexity involved in the annotation, there are very few public multiplex imaging datasets that contain reference annotations. The complexity of the annotation is especially clear when it comes to annotating interactions between cell phenotypes or between cellular neighborhoods.

Synthetic images can be used for benchmarking image analysis algorithms or used as an part of training datasets in machine-learning frameworks [22][23]. Several synthetic benchmarks have been developed to simulate cells with subcellular resolution acquired by optical and fluorescence microscopy. Filogen [24], for instance, is a simulator capable of generating 3D time-lapse sequences of motile cells with filopodial protrusions acquired by a confocal fluorescence microscope. It has been used to create synthetic images with reference annotations for the development of cell tracking algorithms [25]. Alternatively, generative adversarial networks (GANs) have been trained with annotated real fluorescence microscopy data to create realistic 3D stacks of fluorescent cells [26]. However, to the best of our knowledge there are not simulators of tissue sections replicating the complex interactions of tumor microenvironment elements *in situ*. Here, we present **Synplex**, a **Syn**thetic multi**plex** tissue simulator that mimicks multiplex immunostained cancer tissues, in which cells lay on a bidimensional plane following user-defined cell-level and neighborhood (i.e. cell interaction)-level rules. These rules can define specific disease paradigms, from the number of markers and cell phenotypes present in the tissue, the number and composition of cellular neighborhoods, cell phenotypes interactions and neighborhood interactions. The output of our simulator consists of sets of multiplex images and their corresponding ground truths, where masks identify regions with specific TMEs. The image generation process follows an iterative optimization process starting from a random initialization that introduces a realistic variability between images. This way, image sets of different patient-type cohorts can be created with certain intra-class variability, as it happens in real patient datasets. The proposed simulation system is validated by generating a specific disease paradigm, inspired in a real study. We show that the generated images can be useful to qualitatively and quantitatively assess computational solutions that address different tasks, such as tissue segmentation, cell segmentation, or patient classification. **Synplex** has been designed bearing especially in mind tissue events that take place in tumor development, but its flexibility makes it suitable to create synthetic scenarios reflecting other diseases.

## 2  Methods

The simulation of multiplex stained tissues consists of three sequential steps (**Fig. 1**): modeling cellular neighborhoods, modeling cell phenotypes, and finally the generation of a realistic tissue texture. In our synthetic simulator, the tissue is represented by two overlapping ground-truth masks, where cell phenotypes lay in specific cellular neighborhoods. Each steps is carried out, sequentially, by a different software component: i.e. the cellular neighborhood module, cell phenotype module, and texture module, described in the following sections.

### 2.1  Modeling of cellular neighborhoods (Algorithm 1).

The user must define four parameters: i) the number of cellular neighborhoods $N$: ii) a vector containing the relative abundance of each neighborhood in the tissue $N_{ab} \in \mathbb{R}^N$, where the sum of its values represents the whole tissue i.e., $N_{ab}(1) + N_{ab}(2) + \ldots + N_{ab}(N) = 100\%$. iii) a matrix that reflects the pairwise interactions between neighborhoods $N_{int} \in \mathbb{R}^{N \times N}$, which can be positive (attraction) negative (repulsion), or zero (no interaction between elements). iv) the dimensions of the multiplex image $m_x, m_y$ in the x and y-axis, respectively.

The process starts by creating and initializing a neighborhood image mask $M_{Nb} \in [0..N]^{m_x \times m_y}$ with random pixel values between 1 and $N$ (**Fig. 2a, d**). Then an index image $R_{Nb} \in [0,1]^{m_x \times m_y}$ is initialized with ones. This image will be used to keep track of the pixels that have already been permanently assigned to a neighborhood by the tissue simulator. Once initialized, the mask values of $M_{Nb}$ are updated iteratively using the following constrained optimization function that seeks to minimize $\ell_N$:

$$\ell_N = sum\left( abs\left( N_{int} \odot \left(\frac{N_{ab}}{M_{Nb_{ab}}}\right)^2 \right) \right) \tag{1}$$

where $M_{Nb_{ab}} \in \mathbb{Z}^N$ is a vector that contains the actual neighborhood abundance present in $M_{Nb}$, and $N_{int}$ constrains the function loss to specific neighborhood interactions. Note that the $\odot$ symbol denotes elementwise multiplication, and $sum$ stands for a column-wise sum operation. The concept behind this optimization rule is, given a random neighborhood mask $M_{Nb}$, to find the optimal replacement of local neighborhoods that may eventually create a user-defined neighborhood interaction with specific abundances in the tissue.

**Algorithm 1**. Pseudocode for modeling cellular neighborhoods. **Inputs**: $N_{ab}$, Neighborhood abundance matrix. $N_{int}$, Neighborhood to neighborhood interaction matrix, $\{m_x, m_y\}$ tissue image size in x and y-axis, respectively. $N$ number of neighborhoods. **Output**: $M_{Nb}$, cellular neighborhood mask.

| | |
|---|---|
| 1: | Random neighborhood mask: $M_{Nb} = randomMatrix(m_x, m_y, N)$ |
| 2: | Initialize vector of *set* indices: $R_{Nb} = matrixOfOnes(m_x, m_y)$ |
| 3: | **while** $\sum R_{Nb} > 0$ |
| 4: |    Update optimization rule: $U = N_{int} \odot \left(\frac{N_{ab}}{M_{Nb_{ab}}}\right)^2$ |
| 5: |    Select one random pixel $s \mid s \in R_{Nb} = 1$ |
| 6: |    Get context information of $s$: $M_C = \{M_N \mid M_N: dist(M_N, s) < 12 \text{ pixels}\}$ |
| 7: |    **if** $max(histcounts(M_C)) > 90\%$ **then** |
| 8: |      Most abundant neighborhood: $M_{Nb} = argmax(histcounts(M_C)): M_N \cap M_C$ |
| 9: |      Reduce pool of pixels to iterate: $R_{Nb} = 0: R \cap M_C$ |
| 10: |    **Else** |
| 11: |      Update set of neighborhoods: $M_{Nb} = U(M_C): M_{Nb} \cap M_C$ |
| 12: | **end while** |

In each iteration of the optimization, the tissue simulator optimizes a 24x24 pixel region surrounding a randomly selected pixel $s_{i,j}$. This pixel is randomly selected among those that have not already been assigned to a specific neighborhood, i.e. $R_{Nb}(i,j) = 1$. To this end, a matrix $M_c \in \mathbb{R}^{24 \times 24}$ is created with the values of $M_{Nb}$ contained within that 24x24 region. This matrix is used to update their corresponding values in $M_{Nb}$. Two strategies are followed, depending on the values of $M_c$:

- (i) If >90% pixels of $M_c$ correspond to a certain neighborhood, the simulator automatically assigns $s_{i,j}$ to that neighborhood, following a majority voting strategy. Then, as the pixel value $s_{i,j}$ has been *set*, the simulator sets $R_{Nb}(i,j)$ to zero, which means that it will not be chosen by the simulator again and the assigned value is permanent.
- (ii) Otherwise, the simulator updates each value of $M_c$ using the update rule matrix $U = N_{int} \odot \left(\frac{N_{ab}}{M_{Nb_{ab}}}\right)^2$. To this end each element $M_c(n,m) = v$ is replaced by the index of the maximum of the $v^{th}$ row of U. This way, pixel values are optimally transformed in order to minimize $\ell_N$, while accounting for $N_{int}$ and $N_{ab}$.

This iterative process continues until all pixel values are permanently assigned by the tissue simulator i.e. $\sum R_{Nb} = 0$. When the process stops, the resulting $M_{Nb}$ neighborhood mask has N interacting neighborhoods following the rules specified by the user (**Fig. 2i**). This mask will be used later by the cell phenotype module.

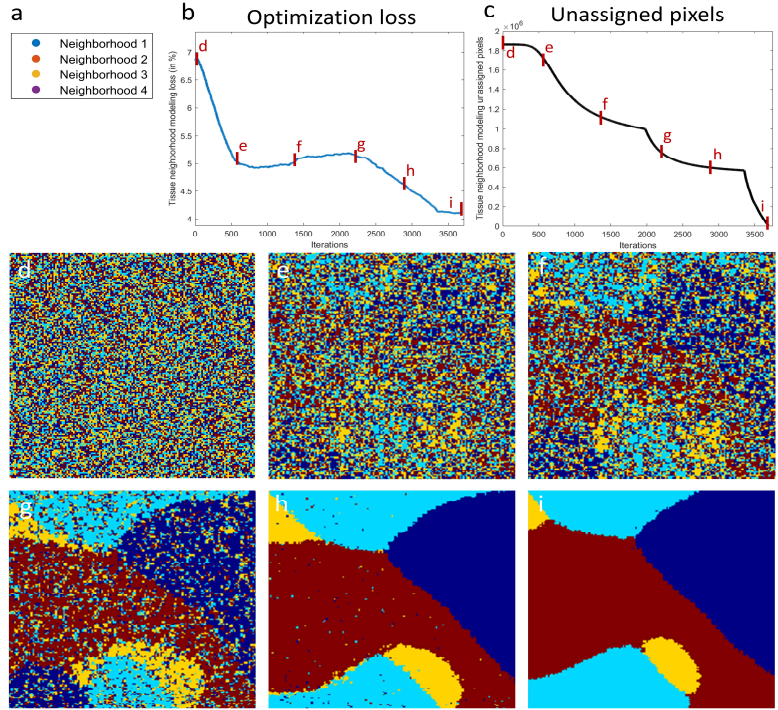

**Figure 2**. Visualization of the tissue generator optimization process to generate a tissue neighborhood ground-truth. **a.** Color legend. **b.** In blue, optimization loss. **c.** Number of unassigned pixels. **d.** Random initialization of the neighborhood mask. **e,f,g,h.** neighborhood mask at 700, 1400, 2100, and 2800 iterations, respectively. **i.** Final optimized neighborhood mask containing $N$ interacting neighborhoods.

## 2.2 Modeling cell phenotypes (Algorithm 2).

The user must define five parameters: i) the number of phenotypes $P$ in the tissue. ii) the relative abundance of each phenotype in each tissue neighborhood $P_{ab} \in \mathbb{R}^{P \times N}$, where each column in the matrix represents the complete neighborhood e.g., $P_{ab}(1,1) + P_{ab}(2,1) + \ldots + P_{ab}(P,1) = 100\%$. iii) the interaction between phenotypes in each neighborhood $P_{int} \in \mathbb{R}^{P \times P \times N}$, where the third dimension contains phenotype interactions occurring within the neighborhood, which can be positive (attraction) negative ( repulsion), or zero (no interaction between cell phenotypes). iv) the eccentricity of cells for each phenotype $P_{ecc} \in \mathbb{R}^P$, ranging from zero (round cells) to 1 (completely elongated cells). v) the size of the cells of each phenotype $P_{siz} \in \mathbb{R}^P$, measured in pixels. Please notice that the previously computed $M_{Nb}$ cellular neighborhoods mask is used to locate cell phenotypes in the corresponding neighborhood. Similar to what was done to model cellular neighborhoods, the process

**Algorithm 2.** Pseudocode for modeling cell phenotypes. **Inputs**: $P_{ab}$, neighborhood abundance matrix. $P_{int}$, cell phenotype to cell phenotype interaction matrix, $P_{ecc}$, eccentricity of cell phenotypes, $P_{siz}$, size of cell phenotypes, $\{m_x, m_y\}$ tissue image size in x and y-axis, respectively, $M_N$, the cellular neighborhood mask, and $P$ the number of phenotypes. **Output**: $M_P$, the cell phenotype mask.

1:     Initialize neighborhood mask: $M_P = randomMatrix(m_x, m_y, P)$
2:     Initialize vector of indices: $R = matrixOfOnes(m_x, m_y)$
3:     **while** $\sum R > 0$
4:        Update optimization rule: $U = P_{int} \odot \left(\frac{P_{ab}}{M_{P_{ab}}}\right)^2$
5:        Select random pixel $s \mid s \in R_s = 1$
6:        Get pixel context information of $s$: $M_C = \{M_P \mid M_P: dist(M_P, s) < 24 \text{ pixels}\}$
7:        Obtain neighborhood: $Ng = majorityVoting(\{M_N \mid M_N: dist(M_N, s) < 24 \text{ pixels}\})$
8:        Obtain cell phenotype: $Ph = majorityVoting(U'(M_C)) \mid U \xrightarrow{Ng} U'$
9:        Generate cell phenotype mask: $M_C' = generateEllipse(s, P_{ecc}, P_{siz}, Ng)$
10:     **if** $mean(R \cap M_C') > 80\%$ **then**
11:        Set new cell phenotype: $M_P = M_C': M_P \cap M_C'$
12:        Reduce pool of pixels to iterate: $R = 0: R \cap M_C'$
13:     **elseif** $mean(R \cap M_C') < 20\%$ **then**
14:        Set background phenotype: $M_P = Background : M_P \cap M_C'$
15:        Reduce pool of pixels to iterate: $R = 0: R \cap M_C'$
16:     **else**
17:        Set new cell phenotype: $M_P = M_C': M_P \cap M_C'$
18:     **end while**

starts by creating a cell phenotype image mask $M_{Pn} \in [0..P]^{m_x \times m_y}$ in which each pixel has initially a random value that specifies a cell phenotype between 1 and $P$ (**Fig. 3a,d**). Additionally, a matrix of indices $R_{Pn} \in [0,1]^{m_x \times m_y}$ is initialized to ones, and it is used to control whether each pixel of the image has already been permanently set to a cell phenotype or it still needs to be assigned by the tissue simulator. Next, these randomized mask values ($M_{Pn}$) are updated iteratively using the following constrained optimization problem that seeks to minimize $\ell_P$:

$$\ell_P = sum\left(abs\left(P_{int} \odot \left(\frac{P_{ab}}{M_{Pn_{ab}}}\right)^2\right)\right) \quad (2)$$

where $P_{ab}$ is the objective phenotype abundance user-defined parameter, $M_{Pn_{ab}} \in \mathbb{R}^{P \times N}$ is the actual neighborhood abundance present in $M_{Pn}$, and $P_{int}$ is the matrix that constrains the function loss to specific cell phenotype interactions. Note that the $\odot$ symbol denotes elementwise multiplication, and $sum$ stands for 3 sequential column-wise sum operations. As done in the previous subsection, to minimize $\ell_P$, the tissue simulator optimizes a subset of pixel values in each iteration, and subsequently updates $M_{Pn_{ab}}$. We use $U = P_{int} \odot \left(\frac{P_{ab}}{M_{Pn_{ab}}}\right)^2 \in \mathbb{R}^{P \times P \times N}$ as the update rule to minimize $\ell_P$.

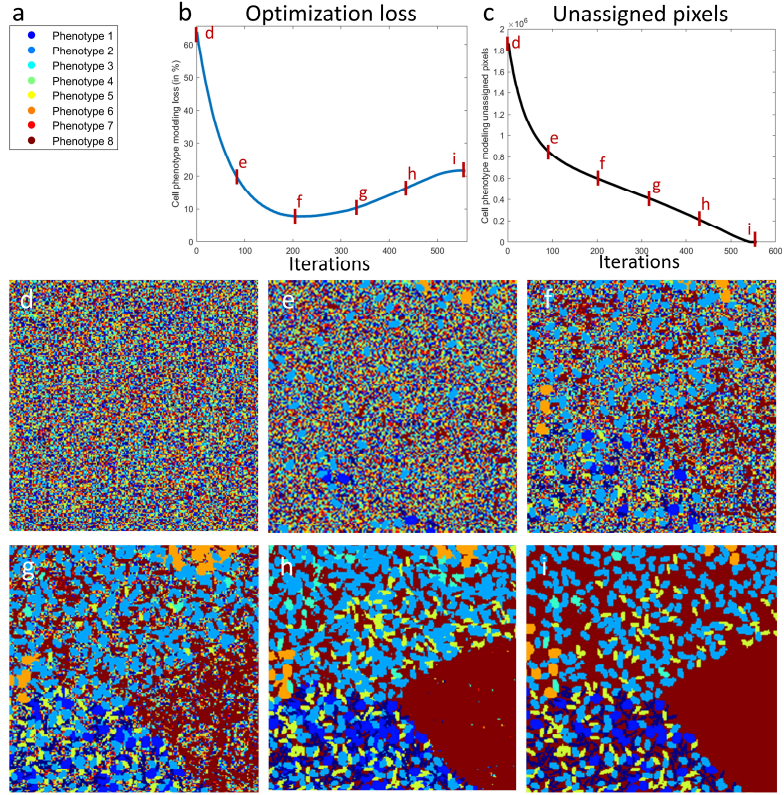

**Figure. 3**. Visualization of the cell phenotype optimization process to generate a cell phenotype ground-truth. **a.** Color legend. **b.** In blue, optimization loss. **c.** Number of unassigned pixels. **d.** Random initialization of the cell phenotype mask. **e,f,g,h.** Cell phenotype mask at 110, 220, 330, and 440 iterations, respectively. **i.** Final optimized cell phenotype mask containing $P$ interacting cell phenotypes.

In each iteration, the optimization process selects a region of 24x24 pixels from $M_{Pn}$, surrounding a randomly selected pixel $s_{i,j}$, with the only condition that it has not already been *set* to a specific cell phenotype, i.e. $R_{Pn}(i,j) = 1$. The phenotypes of this region extracted from $M_{Pn}$ are stored in an auxiliary 24x24 matrix $M_{C1}$.

The neighborhoods of the corresponding 24x24 region, extracted from $M_{Nb}$ are stored in another auxiliary 24x24 matrix $M_{C2}$. To assign the cell phenotype of the corresponding pixel $s(i,j)$ two steps are followed: First, we use a majority voting strategy on $M_{C2}$ to determine the most frequent neighborhood ($Ng$) contained in the region surrounding pixel $s(i,j)$.

Then the simulator identifies the new value of the phenotype that should be assigned to all pixels in $M_{C1}$ using the rule matrix $= P_{int} \odot \left(\frac{P_{ab}}{M_{Pn_{ab}}}\right)^2 \in \mathbb{R}^{P \times P \times N}$. This is done in two steps: first we extract the $Ng^{th}$ plane of U, and create a new matrix $U'$. Then for all values of $M_{C1}$ (e.g. v, z, …) we add the rows of $U'$ that correspond to their values (i.e. we sum the v$^{th}$ row + z$^{th}$ row + … ) and then store the index of the resulting vector that contains the maximum value $Ph$. Once this optimal cell phenotype has been selected, we create a new binary 24x24 cell mask $M_{C1}'$ centered in $s(i,j)$ using an ellipse generator with the following parameters for the corresponding phenotype: $P_{ecc}$ measured as the ratio of major to minor axis, and $P_{siz}$ as the size of the major axis.

Finally, to decide whether to *fix* $M_{C1}'$ values to $M_{Pn}$ three options are considered, based on the distribution of values of $R_{pn}$ in the corresponding 24x24 region:

- (i) If more than 80% of $M_{C1}'$ values have not been *set* yet, i.e. $mean(R_{pn} \cap M_{C1}') > 80\%$, the simulator automatically assigns $M_{C1}'$ to $R_{pn}$ in that region. Additionally, as the pixel values $M_{C1}'$ have been *fixed*, the simulator sets to zero all the corresponding pixels of $R_{pn}$, which means that they will not be chosen by the simulator again.
- (ii) If less than 20% of $M_{C1}'$ values are yet to be *set*, i.e. $mean(R_{pn} \cap M_{C1}') < 20\%$, the simulator automatically assigns the background phenotype to $M_P$ (Phenotype 8 in **Fig. 3**). Additionally, as the pixel values $M_{C1}'$ have been now *fixed*, the simulator sets the corresponding pixels of $R_{pn}$ to zero, which means they will not be chosen by the simulator again.
- (iii) If none of the previous conditions apply, the simulator automatically assigns $M_{C1}'$ to $M_P$. In this case, $M_{C1}'$ is not *fixed* yet, and therefore $R_{pn}$ values are not changed.

The iterative process continues until all pixel values are permanently assigned to a phenotype, i.e. $\sum R_{pn} = 0$. When the process stops, the resulting $M_{Pn}$ cell phenotype mask has P interacting phenotypes that follow the rules specified by the user.

### 2.3 Tissue textures: simulation of acquisition device

Once the phenotype and neighborhood masks are generated, the combination of markers that defines each phenotype is applied to each pixel, thus generating a $C \times m_x \times m_y$ matrix, where $C$ is the number of markers and, $m_x$ and $m_y$ are the size of the image in the x and y-axis respectively. To create final realistic images, we first artificially simulate the mixing of fluorescent probes, which is caused by overlapping fluorescent emissions and produces signal leakage between consecutive channels [27]. To this end, we apply a gaussian filter across the z-dimension, thus leaking some of the marker's intensity into the adjacent channels. Then, we simulate the blurring effect of a microscope's point spread function by applying a gaussian filter with a standard deviation of 0.75 to the image. Finally, we simulate the dark current noise by adding gaussian noise to obtain an SNR of 20 dB.

## 3 Experiments and results

To demonstrate the capability and flexibility of the proposed **Synplex** we generated a set of 30 images containing the tumor microenvironments of a unique predefined disease model. The following paragraphs describe the set of user-defined parameters and the qualitative and quantitative evaluation of the generated TMEs.

### 3.1    Simulation of tissue samples

We used **Synplex** to create 30 1000x2000x6 synthetic tissue images that contain 9 cell phenotypes (Ph1-Ph9), defined by the level of expression of 6 markers (Mk1-Mk6) (**Fig. 4a**), where the first six cell phenotypes (Ph1-Ph6) consist of cells stained by only one marker at different expression levels, cell phenotypes Ph7 and Ph8 show cells co-expressing 2 markers, and Ph9 corresponds to the background i.e., no expression of markers. Moreover, each cell phenotype shows a user-defined cell size (**Fig. 4b**), and shape (**Fig. 4c**), where cells range from an elongated morphology (Ph5) to a round one (Ph3), and show different sizes from a radius of only 2 pixels (Ph4) to 6 pixels (Ph2, Ph6). Local interactions between cell phenotypes are also defined (**Fig. 4d**) that give rise to 6 types of neighborhoods (Nb1-Nb6) (**Fig. 4e**). As an example, Nb2 contains local interactions between 4 cell phenotypes: Ph3 and Ph7 have a relative abundance of 20% in the neighborhood and they repel each other (i.e., they tend to be spatially separated); in contrast, Ph4 and Ph5 have a relative abundance of 10% in the neighborhood and they attract each other (i.e., they tend to be spatially close from each other). All neighborhoods have a predefined prevalence in the tissue (**Fig. 4f**) Here, Nb1 corresponds to the background (i.e. absence of cells). Finally, neighborhood to neighborhood interactions can be defined (**Fig. 4g**). In this experiment, two neighborhoods (Nb2 and Nb3) attract each other (i.e., they tend to be spatially close to each other in the image), whereas two other neighborhoods (Nb5 and Nb 6) repel each other (i.e., they tend to be spatially separated).

### 3.2    TME quantitative and qualitative evaluation

**Quantitative Evaluation**

As **Synplex** is initialized from a random mask both at the cell phenotype and cellular neighborhood levels, we expect to find a certain variability between the generated TME elements. This allows mimicking a realistic scenario, where patients' tissues are usually inhomogeneous and there are variabilities inherent to the multiplex imaging technique as well.

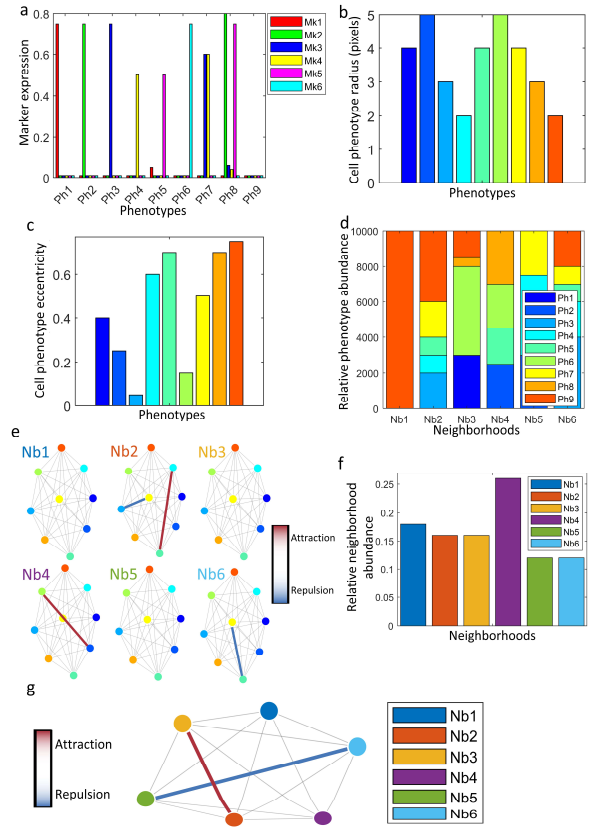

**Figure 4**. Set of parameters used for the experiment in the synthetic simulator. **a.** Relative marker expression level for each phenotype colored by marker. Note that the color shown in the legend for each marker is the one used to display the multiplex image. **b.** Cell phenotype size (cell radius) in pixels **c.** Cell phenotype eccentricity measured as the ratio of major to minor axis. **d.** Relative phenotype abundance for the six different neighborhoods **e.** Cell to cell interaction between cell phenotypes, where a red line represents an attraction between two of them, and a blue line a repulsion. The rest of cell phenotypes show no interaction among them. The colormap representing the level of attraction/repulsion is also shown. **f.** Relative neighborhood abundance in the tissue. **g.** Neighborhood to neighborhood interactions, together with the colormap that codifies the level of attraction/repulsion between any two of them.

We first analyzed the 30 ground-truth cellular neighborhood masks generated by **Synplex** and studied their simulated neighborhood interactivity. To this end, we extracted neighborhood contours (i.e. neighborhood-neighborhood adjacency) and calculated the number of contour pixels that each pair of neighborhoods share. Here, a high level of neighborhood-neighborhood adjacency (i.e., large number of contour pixels shared) is considered as a high level of attraction, whereas a low level of adjacency is interpreted as repulsion between pairs of neighborhoods. In **Fig. 5a,b,c** we show the user-defined matrix of neighborhood interactions (**Fig. 5a**), which corresponds to what is graphically shown in **Fig. 4e**. A boxplot of the number of contour pixels shared between Nb2 and each of the other neighborhoods (**Fig. 5b**) reflects that Nb2 and Nb3 have a high attraction, whereas a boxplot of the number of contour pixels shared between Nb5 and each of the other neighborhoods (**Fig. 5c**) shows a lower relative interaction and, therefore, a correctly simulated repulsion between Nb5 and Nb6.

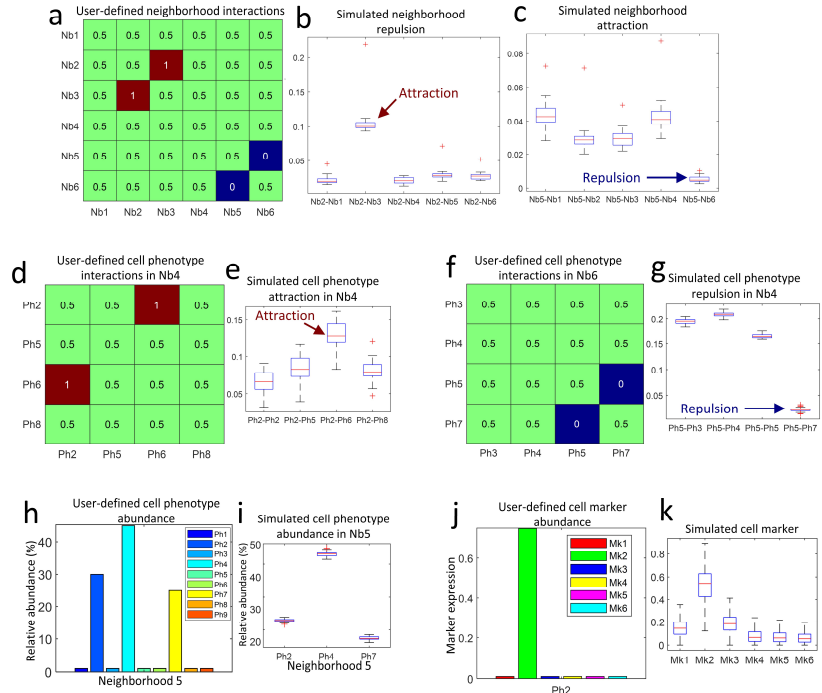

**Figure 5**: Quantitative evaluation of the 30 simulated tumor microenvironments. **a.** User-defined cellular neighborhood interactions, **b.** simulated repulsion and **c.** simulated attraction. **d,f.** User-defined cell phenotype interactions in Nb4 and Nb6, **e.** simulated phenotype attraction in Nb4, and **g.** simulated phenotype repulsion in Nb6. **h.** User-defined cell phenotype abundance for Nb5, **i.** actual cell phenotype abundance for Nb5 measured in the generated images. **j.** User-defined cell marker abundance for Ph2, **k.** actual cell marker abundance for Ph2 measured in the generated images.

Then we analyzed the 30 ground-truth cell phenotype masks, and studied their interactivity for each neighborhood, their prevalence in cellular neighborhoods, and the markers' expression for each specific cell phenotype. In this case, the interaction between cell phenotypes is calculated from a network of interconnected cells, where each cell is connected to other neighboring cells located within a radius of 12 pixels. Here, the interaction between cell phenotypes is measured as the number of 1-hop connected cells of each pair of phenotypes, where a low number reflects that few cells of both phenotypes are found within a context of 12 pixels, thus those phenotypes repel each other, and a high number shows an attraction between them. In **Fig. 5d,e**, Nb4 is

13composed by 4 cell phenotypes (Ph2, Ph5, Ph6, and Ph8), where Ph6 and Ph2 are attracted to each other. **Fig. 5e** shows a boxplot where, in accordance with the configuration of the matrix of phenotype interactions for Nb4 (**Fig. 5d**), **Synplex** effectively simulated an attraction between Ph6 and Ph2. Moreover, we show in **Fig. 5f,e** the phenotype interactions found in Nb6, where the repulsion between Ph5 and Ph7 configured in the matrix of neighborhood interactions was indeed apparent in the simulated images. Next, we studied whether the abundance of cell phenotypes found in the images correctly followed the ones that were defined in the configuration setup (**Fig. 5h**). **Fig. 5i** shows that **Synplex** simulated a correct relative abundance for Ph2, Ph4, Ph7 in Nb5 with a low standard deviation.

Finally, we evaluated the final multispectral images generated by the tissue texture module. Here, the generated cell phenotype mask and a matrix showing the cell marker expression of cell phenotypes (e.g. user-defined Ph2 markers' expression in **Fig. 5j**) were input to the module simulating two effects: the blurring effect of a microscope's point spread function, and the mixing of overlapping fluorescent probes resulting in signal leakage between consecutive spectral channels and, thus, markers. The first effect can be observed in **Fig. 5k** as the marker expression (Mk2) standard deviation of Ph2 cells, which can be controlled with the chosen SNR and the deviation of the gaussian filter. The second effect can be seen in **Fig. 5k**, where the expression of Mk1 and Mk3 for Ph2 is increased when compared with the configuration setup (**Fig. 5j**) as a result of the leakage of Mk2 in their respective spectral channels. In other words, Ph2 was configured as cells that only express Mk2, but the simulated spectral mixing effect creates a realistic image where the signals acquired for Ph2 cells in Mk1 and Mk3 spectral channels contain a certain level of expression due to the leakage from the Mk2 channel.

**Qualitative Evaluation**
We qualitatively evaluate the generated multiplex images that were produced by the tissue texture module. To this end, we first show a set of 30 simulated multiplex images (**Fig. 6a**), showing tumor microenvironments that follows a unique disease paradigm. Then, a crop from multiplex image 1 was extracted to zoom in on the structures present in the tissue. In **Fig. 6b**, we can visualize differences between the image background and unstained tissue i.e., negative cells for Mk1. This is caused by the spectral leakage that was simulated by inserting marker expression from consecutive markers. In **Fig. 6c**, we show tissue elements stained by Mk2 and Mk5, showing visually, an attraction between cells expressing $Mk2^+$, and $Mk2^+Mk5^+$.

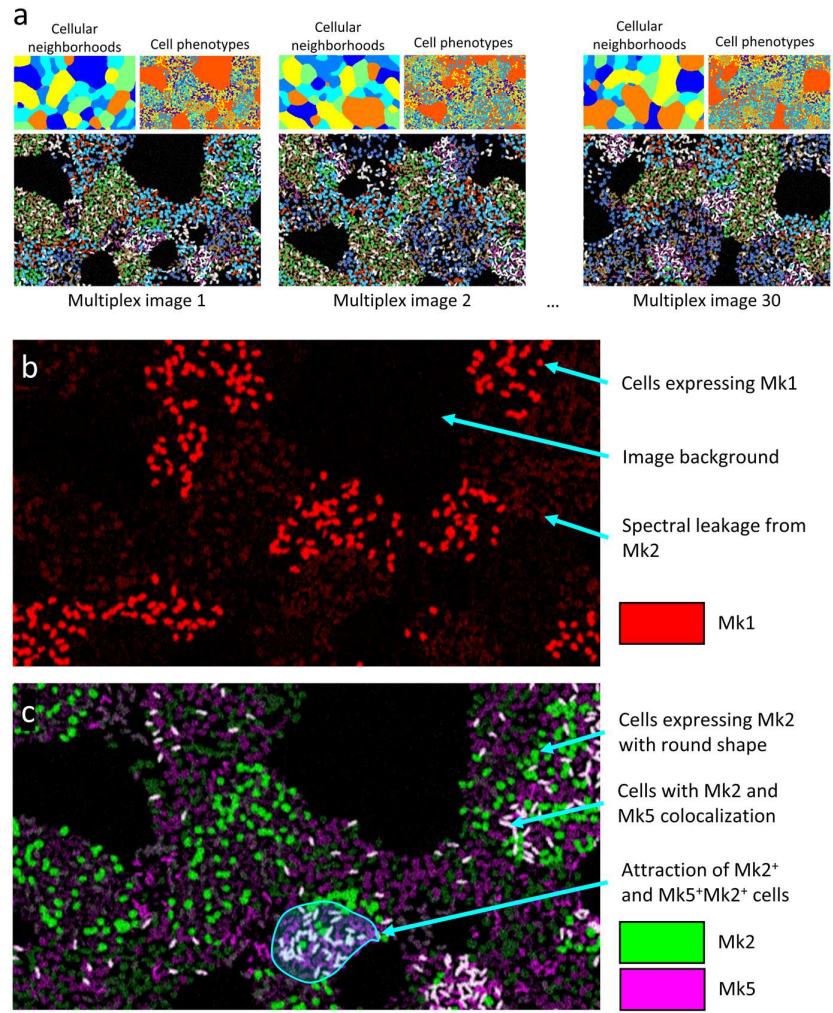

**Figure 6**: Qualitative assessment of the generated synthetic tissues. **a.** Set of multiplex images with their inherent generated ground-truth masks. **b.** Tumor microenvironment elements stained by marker 1 considering technical variabilities. **c.** Tumor microenvironment elements stained by marker 2 and 5, showing a cellular attraction.

## 4   Discussion

We have presented a novel simulator, called **Synplex**, that generates synthetic tumor multiplex images representing tumor microenvironments that contain realistic and adjustable interactions between cell phenotypes and cellular neighborhoods. **Synplex** is composed of three modules: i. generator of a ground-truth mask of interacting cellular neighborhoods with a specific prevalence in the tissue, ii. generator of a ground-truth mask of interacting cell phenotypes with specific prevalence in each neighborhood, and iii. generator of multiplex images displaying technical variabilities inherent to the multispectral acquisition process.

**Synplex** was validated by comparing the statistical values extracted from the simulated ground-truth masks and the original user-defined parameters. This strategy allowed us to show the accuracy of **Synplex** to simulate the specific expression of markers, cell phenotype abundances in neighborhoods, cell phenotype interactions, and neighborhood interactions.

The fact that our simulator automatically creates completely annotated bioimage data may stimulate the development of new image analysis pipelines to analyze these types of images. Among others, these artificial datasets could be useful for the augmentation of annotated datasets destined to train machine learning pipelines, and to create datasets for benchmarking novel tissue analysis algorithms. Indeed, the annotation of thousands of cells in multiplex immunostained tissues is often impracticable, not allowing researchers to train machine learning models to perform cell segmentation, tissue segmentation, or patient classification. To this end, we believe **Synplex** will be of great value by generating realistic tissues, may it be for the classification of interacting cells, or the segmentation of cells with tunable fluorescent levels. Furthermore, benchmarking new bioimage analysis tools is often a complicated task due to the scarcity of annotated data, which is even more complex in the case of multiplex imaging data. Here, the interactivity between cell phenotypes or neighborhoods is often weak, and thus, tools that aim to make relevant biomedical discoveries by quantifying those relationships must be systematically validated. To this end, **Synplex** allows researchers to create images with adjustable parameters, generating, for instance, artificial patient cohorts, where patient types can be defined by specific variations in elements from the tumor microenvironment. Using **Synplex**, it is possible to tune how present is the phenotype in the whole tissue, thus creating different levels of difficulty for the analysis tool, thus validating algorithms that aim to find clinically relevant elements in the image.

All in all, thanks to the stochastic nature of **Synplex** every tissue image is generated independently. This allows the generation of batteries of images that can be used for very different purposes and will help researchers to systematically generate the next generation of bioimage analysis tools.

# 5 Acknowledgements

This work was funded by the Spanish Ministry of Science, Innovation and Universities, under grants number RTI2018-094494-B-C22 and RTC-2017-6218-1 (MCIU/AEI/FEDER, UE)